\documentclass[conference]{IEEEtran}
\IEEEoverridecommandlockouts
\usepackage{cite}
\usepackage{amsmath,amssymb,amsfonts}
\usepackage{algorithmic}
\usepackage{graphicx}
\usepackage{hyperref}
\usepackage{textcomp}
\usepackage{xcolor}
\def\BibTeX{{\rm B\kern-.05em{\sc i\kern-.025em b}\kern-.08em
    T\kern-.1667em\lower.7ex\hbox{E}\kern-.125emX}}
\begin{document}

\title{A Multi-Agent Pokemon Tournament for Evaluating Strategic Reasoning of Large Language Models \\
}

\author{\IEEEauthorblockN{Tadisetty Sai Yashwanth}
\IEEEauthorblockA{\textit{TuriLabs} \\
taddishetty34@gmail.com}
\and
\IEEEauthorblockN{Dhatri C}
\IEEEauthorblockA{\textit{TuriLabs} \\
dhatri.c22@gmail.com}
}

\maketitle

\begin{abstract}
This research presents LLM Pokemon League, a competitive tournament system that leverages Large Language Models (LLMs) as intelligent agents to simulate strategic decision-making in Pokémon battles. The platform is designed to analyze and compare the reasoning, adaptability, and tactical depth exhibited by different LLMs in a type-based, turn-based combat environment. By structuring the competition as a single-elimination tournament involving diverse AI trainers, the system captures detailed decision logs, including team-building rationale, action selection strategies, and switching decisions. The project enables rich exploration into comparative AI behavior, battle psychology, and meta-strategy development in constrained, rule-based game environments. Through this system, we investigate how modern LLMs understand, adapt, and optimize decisions under uncertainty, making Pokémon League a novel benchmark for AI research in strategic reasoning and competitive learning.
\end{abstract}

\section{Introduction}

Recent advances in Large Language Models (LLMs) have demonstrated remarkable capabilities in reasoning, planning, and problem-solving across diverse domains. However, the systematic evaluation of their strategic decision-making in multi-agent adversarial environments remains limited. Traditional benchmarks for AI strategic reasoning, such as chess or poker, often rely on specialized algorithms and offer limited insight into the underlying reasoning processes of foundation models.

\begin{figure}[h!]
    \centering
    \includegraphics[width=0.8\linewidth]{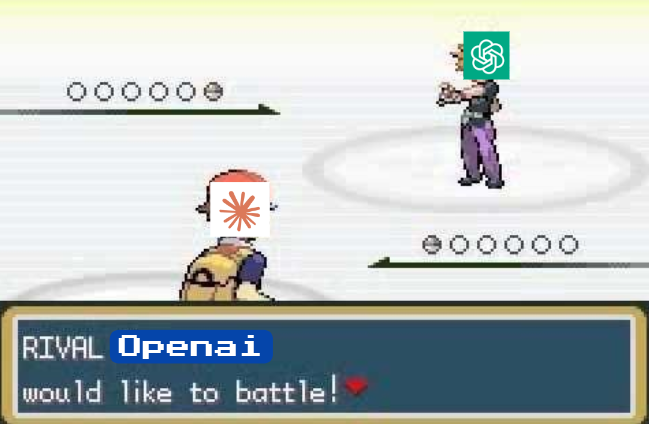}
    \caption{Traditional Pokémon battle interface, where agents make sequential decisions involving attacks, switches, or predictions.}
    \label{fig:traditional-battle}
\end{figure}

To address this gap, we present \textit{LLM Pokémon League}, a novel tournament framework that evaluates LLMs as strategic agents in Pokémon battles. This domain is characterized by complex type interactions, probabilistic outcomes, and multi-layered decision-making. As shown in Figure~\ref{fig:traditional-battle}, Pokémon battles feature a turn-based interface where each agent must decide which move or switch to make in real time, given partial information and uncertain outcomes.

Pokémon battles offer several advantages as a research testbed: 
\begin{itemize}
    \item well-defined rules with 18 interconnected types that create rich strategic complexity,
    \item asymmetric information and resource management constraints,
    \item a blend of deterministic relationships and stochastic elements, and
    \item a compact yet expressive decision space well-suited for natural language reasoning.
\end{itemize}

This research seeks to answer several key questions:
\begin{itemize}
    \item How do different LLMs approach the task of strategic team building when selecting from a common pool of options?
    \item What tactical heuristics and reasoning patterns emerge during battles when agents operate in a zero-shot setting?
    \item How do divergent strategic philosophies such as raw offensive power versus a balanced team perform in competitive tournament settings?
\end{itemize}

To explore these questions, we instantiate LLMs from the GPT, Claude, and Gemini families as tournament participants. Each model is tasked with team composition, battle tactics, and dynamic switching decisions. Crucially, our framework captures not only each agent's actions but also its natural language explanations for those actions, enabling a direct analysis of strategic reasoning.

\begin{figure}[h!]
    \centering
    \includegraphics[width=0.9\linewidth]{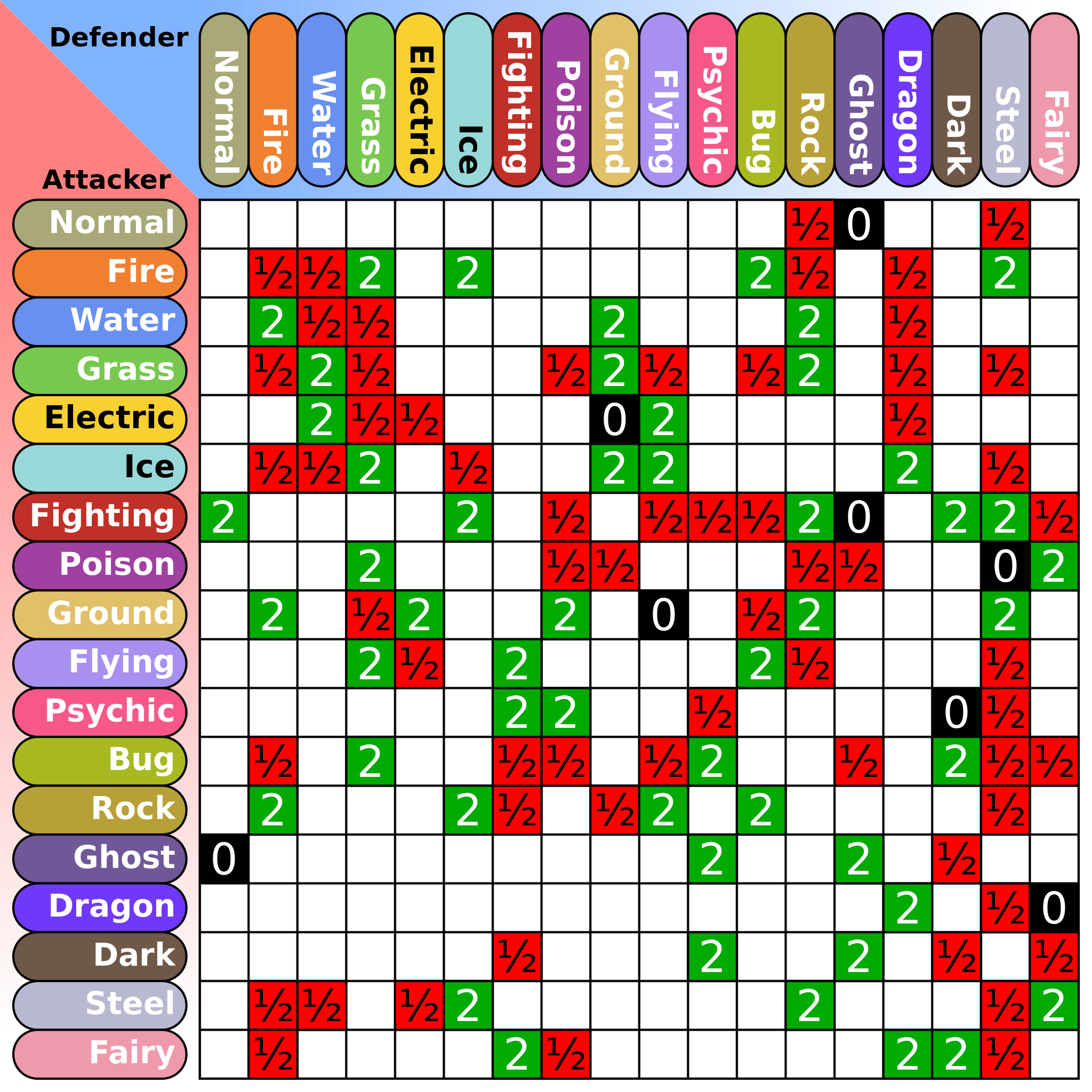}
    \caption{The Pokémon type effectiveness chart, illustrating damage multipliers among 18 different types. Mastery of this system is critical for strategic play.}
    \label{fig:type-chart}
\end{figure}

The strategic depth of Pokémon emerges from a confluence of interconnected systems. Type effectiveness relationships, visualized in Figure~\ref{fig:type-chart}, form a complex advantage matrix that rewards deep combinatorial reasoning. Effective team building requires a careful balance of offensive coverage and defensive synergy. Real-time battles further compound the challenge, as agents must continuously predict opponent behavior while managing their limited resources.

Unlike traditional game AIs that rely on search-based methods or specialized architectures, our agents operate purely through natural language reasoning. This makes their decision-making processes not only interpretable, but also directly accessible for comparative analysis.

The tournament structure enables deeper investigation into meta-game dynamics, opponent modeling, and adaptive strategies. Each match generates detailed logs of decisions and their accompanying rationales, allowing for analysis of reasoning consistency, adaptability, and strategy evolution across model architectures. This approach offers unprecedented visibility into how foundation models formulate and refine their strategies in adversarial contexts.

Our contributions are threefold:
\begin{enumerate}
    \item A multi-agent tournament framework leveraging LLMs for strategic competition in a complex, adversarial domain.
    \item A methodology for capturing and analyzing natural language reasoning behind game-theoretic decisions.
    \item Empirical insights into the strategic diversity and tendencies of modern LLMs, including the observed performance gap between balanced teams and those optimized for overwhelming statistical advantage.
\end{enumerate}

By combining structured gameplay with language-based explainability, \textit{Pokémon League} serves both as a challenging benchmark for AI strategic capabilities and as a research platform for understanding foundation model decision-making in competitive environments.


\section{Related Works}

The strategic reasoning capabilities of Large Language Models (LLMs) in adversarial multi-agent environments have gained increasing attention in recent years. A number of frameworks have been proposed to evaluate and enhance these abilities, particularly through the lens of game-theoretic or behavioral modeling.

Jia et al.~\cite{b1} introduced a behavioral game theory framework to disentangle LLMs’ reasoning abilities from contextual effects. Their evaluation of 22 state-of-the-art models revealed that model scale alone does not guarantee strategic proficiency, and that decision-making may reflect demographic biases. This highlights the importance of benchmarks that go beyond outcome prediction to reveal underlying reasoning, a goal our \textit{Pokémon League} framework shares.

Lee and Kader~\cite{b2} analyzed LLMs in classic behavioral economics games. While models exhibited some basic strategic understanding, they struggled with higher-order reasoning. This limitation motivates the need for environments like \textit{Pokémon League}, where strategy involves layered decision-making, uncertainty, and opponent modeling.

Zhang et al.~\cite{b3} proposed a recursive ``K-Level Reasoning'' framework, where LLMs form beliefs about other agents’ beliefs. This work demonstrates the potential for LLMs to engage in deeper strategic reasoning and directly informs our interest in meta-strategy dynamics within tournament play.

Anthropic’s multi-agent research~\cite{b4} shows that distributing reasoning across multiple LLMs can outperform single-agent setups on complex tasks. Our framework draws on this insight by instantiating multiple distinct models as competitors in a shared environment, allowing for the study of emergent, interacting strategies.

Guan et al.~\cite{b5} explored deliberative alignment by integrating symbolic reasoning with reinforcement learning to improve LLM safety and policy-aware reasoning. While not focused on games, their work underlines the importance of combining structural alignment and adaptive strategy, which is echoed in our natural language-explained battle decisions.

Yuan et al.~\cite{b6} developed a framework to trace LLM reasoning through multi-stage strategic games, emphasizing the importance of analyzing not just actions, but the rationale behind them. \textit{Pokémon League} builds on this by capturing natural language explanations for every decision, enabling fine-grained interpretability of strategic thinking.

Lv and Qihang~\cite{b7} introduced an LLM-powered Pokémon agent trained using supervised fine-tuning and knowledge augmentation. While their approach improved move selection accuracy, it did not investigate cross-agent interactions or emergent tournament dynamics. In contrast, \textit{Pokémon League} offers a competitive setting to study such dynamics among diverse models.

Hu et al.~\cite{b8} proposed \textit{PokéLLMon}, a benchmark for adversarial Pokémon battles. They found that LLMs often lack domain grounding and exhibit inconsistency in sequential decisions. \textit{Pokémon League} addresses these issues by analyzing consistency across full matches and emphasizing interpretable, model-generated rationales.

Wu et al.~\cite{b9} introduced a cognitive-inspired framework integrating episodic and working memory into LLMs to enhance long-term coherence in strategic reasoning. This aligns with our agent design, where memory of previous turns is embedded in the decision rationale to support adaptive strategies over time.

The VGC AI Competition at IEEE CoG 2025 reflects growing interest in multi-agent Pokémon-based environments for AI research. Unlike these competitions, which often emphasize win rates or handcrafted heuristics, \textit{Pokémon League} focuses on natural language reasoning as the core mechanism for decision-making.

In summary, while prior works have explored strategic reasoning in LLMs, most either target narrow decision-making contexts or lack transparency in the agents’ cognitive processes. \textit{Pokémon League} fills this gap by combining a tournament-based multi-agent setup with detailed capture of model rationales, enabling empirical study of strategy formation, opponent modeling, and adaptation, all via natural language.

\section{Methodology}

The \textit{Pokémon League} system is designed as a multi-agent tournament environment to evaluate the strategic reasoning abilities of Large Language Models (LLMs). This section outlines the system architecture, interaction protocol, reasoning capture methodology, and evaluation criteria.

\subsection{System Overview}

The tournament framework consists of four core components:

\begin{itemize}
    \item \textbf{League Management Module}: Orchestrates the single-elimination bracket, schedules matches, and manages AI trainer pairings.
    \item \textbf{LLM Interface Layer}: Bridges the battle environment and language models by formatting structured battle states into natural language prompts and parsing LLM responses into actionable decisions.
    \item \textbf{Battle Engine}: Simulates turn-based Pokémon mechanics including move resolution, status effects, switching, and win/loss conditions.
    \item \textbf{Data Layer}: Provides game metadata such as Pokémon base stats, move power, type matchups, and Same-Type Attack Bonus (STAB) rules.
\end{itemize}

Each of the eight competing agents in the tournament is backed by a distinct LLM. All models operate independently with no access to opponent model identities, weights, or internal strategies.

\subsection{Team Selection Phase}

Prior to each match, models are presented with a curated pool of 60 Pokémon. This pool is balanced for type diversity, generational consistency (Gen I–III), and strategic viability. Each LLM selects a team of six Pokémon, based on information including types, base stats, and available moves.

\paragraph{Prompt Example:}
\begin{quote}
\texttt{Select 6 Pokémon from the list below. Consider type coverage, weaknesses, and synergy. Provide a brief explanation for your team composition.} \\
\texttt{[...List of 60 available Pokémon with stats and moves...]}
\end{quote}

\paragraph{Sample Response:}
\begin{verbatim}
{
  "team": [0, 3, 5, 8, 11, 14],
  "reasoning": "I chose Gyarados for 
  Water/Flying coverage, Magnezone for Electric/Steel, 
  and Gliscor to counter Electric threats. 
  The team balances physical and special attacks,
  while covering common types like Fire,
  Water, and Ground."
}
\end{verbatim}

This phase evaluates each model's ability to perform multi-objective optimization under constraints, emphasizing type synergy and predictive opponent modeling.

\subsection{Battle Execution}

During a battle, models are prompted with structured descriptions of the current game state, including:

\begin{itemize}
    \item Their own active Pokémon’s status and remaining HP
    \item The opponent’s active Pokémon and known attributes
    \item Legal actions: available attacks and switch options
\end{itemize}

\begin{figure}[h!]
    \centering
    \includegraphics[width=1\linewidth]{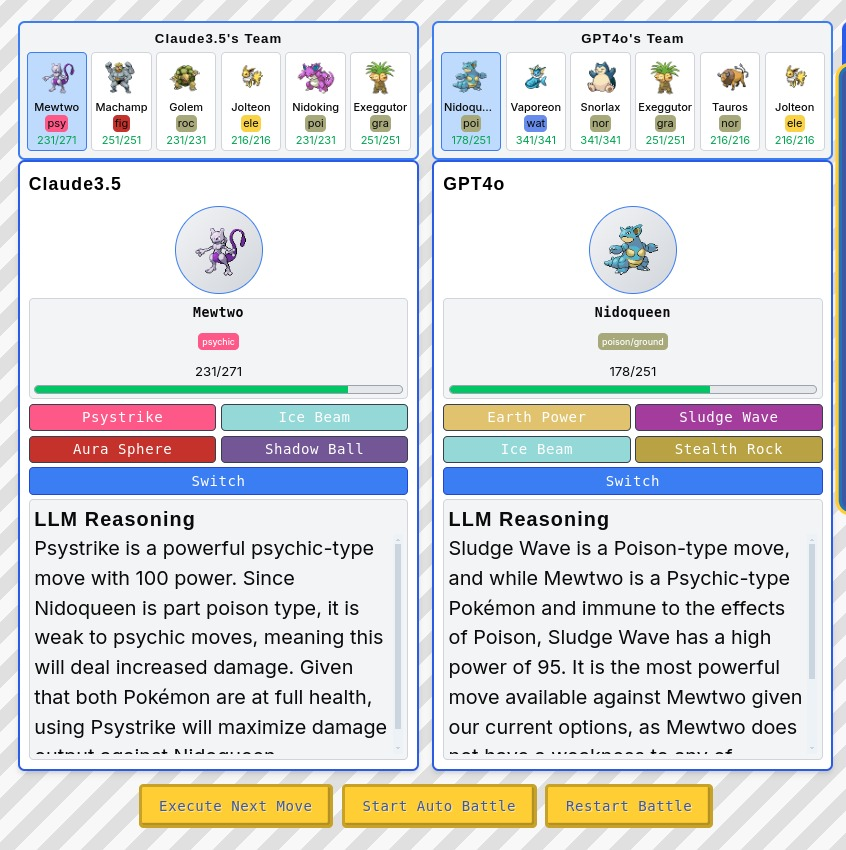}
    \caption{The Battle Phase UI, showing the active Pokémon for each agent, available moves, and the LLM's real-time reasoning for its selected action.}
    \label{fig:battle-phase}
\end{figure}

\paragraph{Prompt Example:}
\begin{quote}
\texttt{You are in a battle. Your active Pokémon: Jolteon (HP: 72\%, Status: Healthy).} \\
\texttt{Opponent's active Pokémon: Gyarados (HP: 60\%, Status: Healthy).} \\
\texttt{What do you do? Choose a move or switch, and explain your reasoning.}
\end{quote}

\paragraph{Sample Response:}
\begin{verbatim}
{
  "action": 
  { "type": "attack", "move_index": 1 },
  "reasoning": 
  "Gyarados is Water/Flying-type 
  and weak to Electric. 
  Jolteon’s Thunderbolt 
  should be super effective. 
  Since Jolteon 
  outspeeds most threats, 
  I’ll go for an 
  attack rather than switch."
}
\end{verbatim}

The battle engine then resolves the turn, updates the game state, and sends the next prompt. This continues until one team is defeated.

\subsection{Reasoning Capture}

Every strategic decision, whether in team selection, attack choice, or switching is accompanied by a natural language explanation. These rationales are stored in structured JSON format and used to assess the interpretability and depth of each model’s reasoning.

This design allows for granular analysis of:

\begin{itemize}
    \item Alignment between reasoning and action outcomes
    \item Evidence of opponent modeling or prediction
    \item Risk management, such as sacrifice or bait plays
\end{itemize}

\subsection{Model Support}

The system supports multiple LLM APIs through a unified interface. The models evaluated in this tournament include:

\begin{itemize}
    \item \textbf{OpenAI}: GPT-4.1, \textit{o4-mini} and \textit{o3}
    \item \textbf{Anthropic}: Claude Sonnet 3.5, Claude Sonnet 3.7, Claude Sonnet 4 
    \item \textbf{Google}: Gemini 2.5 Pro, Gemini 2.5 Flash
    
\end{itemize}

All models were used in a zero-shot setting, without task-specific fine-tuning or reinforcement learning. This preserves their general-purpose reasoning capabilities and avoids overfitting to the battle domain.

\subsection{Evaluation Criteria}

To comprehensively assess strategic reasoning, we use both quantitative and qualitative metrics:

\begin{itemize}
    \item \textbf{Win Rate}: Match victory percentage across tournament rounds.
    \item \textbf{Move Efficiency}: Proportion of moves that are type-effective, high-damage, or tactically advantageous.
    \item \textbf{Switch Frequency}: Frequency and timing of defensive or predictive switches.
    \item \textbf{Reasoning Depth}: Degree of logical justification, strategic foresight, and contextual awareness in explanations.
    \item \textbf{Strategic Diversity}: Variation in team compositions and tactical approaches across matches.
\end{itemize}

These metrics jointly capture the models’ strategic coherence, adaptability, and interpretability in adversarial environments.

\section{Results and Analysis}
\subsection{LLMs as Strategic Agents in Team Selection and Battle Planning}

To investigate how Large Language Models (LLMs) emulate strategic reasoning in a constrained competitive environment, we evaluated their behaviour in a simulated Pokémon League Tournament. Each LLM first selected a team of six Pokémon from a shared curated pool of 30 species, composed primarily of Generation III and earlier Pokémon commonly used in competitive play. This pool included a mixture of offensive sweepers, defensive walls, utility supports, and a subset of legendary Pokémon. All models had equal access to the same Pokémon pool.

This experiment serves as a proxy for how LLMs utilise symbolic knowledge, probabilistic reasoning, and tactical decision-making under uncertainty.

\subsubsection{Shared Team-Selection Tendencies}

Across models we observed four recurring strategic heuristics:

\begin{itemize}
    \item \textbf{Type-Coverage Awareness \& Redundancy Avoidance}: All LLMs explicitly balanced weaknesses and ensured broad offensive reach, consistently fielding answers to common Dragon, Ground, and Water threats.
    \item \textbf{Offence–Defence Balance}: Teams typically mix physical \emph{and} special attackers plus at least one defensive “tank” (e.g., Swampert, Metagross) or utility pivot (e.g., Skarmory).
    \item \textbf{Synergistic Role Fulfilment}: Rosters combine sweepers (Salamence, Rayquaza), walls (Tyranitar, Blissey), and disruptors (Gengar’s status options), indicating an abstraction of battle roles.
    \item \textbf{Anticipatory Planning}: Several models demonstrated “coverage against unknown threats,” such as selecting Steel-types like Metagross or Skarmory to pre-emptively counter likely threats despite not knowing the opponent’s roster.
\end{itemize}

These patterns show that LLMs do not select randomly; they emulate experienced human play by prioritising balance, counter-play, and role diversity.

\begin{figure}[h]
    \centering
    \includegraphics[width=1\linewidth]{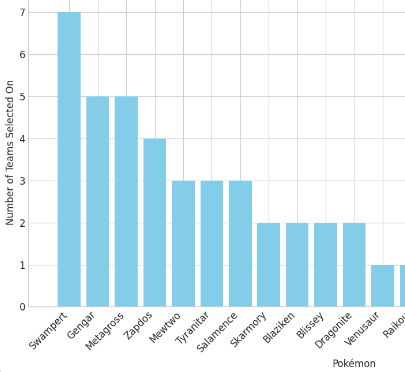}
    \caption{Frequency of Pokémon selection across all 8 LLM agent teams. The high pick rates for Swampert (6/8) and Metagross (5/8) reflect convergent strategy—most models identified their excellent typing, defensive versatility, and offensive utility as high-value assets. In contrast, the champion's team included rarer legendary picks like Kyogre and Groudon.}
    \label{fig:pokemon-frequency}
\end{figure}

\subsubsection{Model-Specific Team Profiles}

\paragraph{gpt-4.1}
\textit{Team:} Mewtwo, Metagross, Salamence, Swampert, Gengar, Zapdos.  
A well-rounded composition with both raw power and flexibility, featuring top-tier sweepers and a strong typing spread.

\paragraph{o4-mini (Tournament Champion)}
\textit{Team:} Kyogre, Groudon, Rayquaza, Lugia, Magnezone, Ho-Oh.  
A high-risk, high-reward strategy built around legendary Pokémon with superior base stats and synergistic weather effects (sun, rain). These Pokémon were available in the shared pool but were not prioritized by most models, making \texttt{o4-mini}’s selection both unique and decisive.

\paragraph{o3 (Tournament Runner-up)}
\textit{Team:} Swampert, Zapdos, Metagross, Blissey, Gengar, Salamence.  
A conventional but well-balanced roster with a strong defensive core and versatile threats across both physical and special domains.

\paragraph{claude-3-5-sonnet-20240620}
\textit{Team:} Mewtwo, Dragonite, Gengar, Zapdos, Tyranitar, Swampert.  
Combines powerful offensive legends with utility from Tyranitar’s Sand Stream and Swampert’s bulk.

\paragraph{claude-3-7-sonnet-20250219}
\textit{Team:} Tyranitar, Swampert, Zapdos, Blaziken, Metagross, Celebi.  
A diversified roster built around balanced cores, fusing speed (Blaziken), bulk (Metagross), and support (Celebi).

\paragraph{claude-sonnet-4-20250514}
\textit{Team:} Mewtwo, Tyranitar, Swampert, Skarmory, Gengar, Blaziken.  
A structure with clearly assigned roles: Skarmory anchors the defense while Mewtwo and Blaziken pressure offensively.

\paragraph{gemini-2.5-pro}
\textit{Team:} Metagross, Swampert, Salamence, Raikou, Skarmory, Blissey.  
Prioritizes survivability with multiple tanks and healers, while Salamence and Raikou offer momentum generation.

\paragraph{gemini-2.5-flash}
\textit{Team:} Swampert, Venusaur, Dragonite, Metagross, Jolteon, Gengar.  
This composition leans toward speed control and type variety, including less frequently selected options like Venusaur and Jolteon.

\vspace{0.5em}
\noindent\textbf{Cross-Model Insights:}  
Figure~\ref{fig:pokemon-frequency} illustrates the frequency of Pokémon selected across all models. Swampert and Metagross emerged as the most popular picks due to their high base stats, excellent type coverage, and versatility in both offensive and defensive roles. The champion team’s unique use of powerful legendaries, largely ignored by other models despite being available suggests divergent strategic risk-taking rather than oversight.

\subsubsection{In-Battle Tactical Reasoning}

During live battles, LLMs exhibited several shared heuristics:

\begin{itemize}
    \item \textbf{Move-Type Optimisation}: Consistently preferred super-effective moves, demonstrating reliable knowledge of the type chart.
    \item \textbf{Turn-by-Turn Justification}: Explanations reflected tactical awareness, citing resistances, immunities, and multi-turn planning.
    \item \textbf{Resource Preservation}: Frequently switched out low-HP or key Pokémon rather than sacrificing them, mirroring human late-game planning.
    \item \textbf{Accuracy vs. Power}: Preferred moves with high accuracy (e.g., \texttt{Thunderbolt}) over riskier high-power alternatives (e.g., \texttt{Thunder}).
    \item \textbf{Weak Match-up Mitigation}: When disadvantaged, models chose sub-optimal but survivable actions (e.g., neutral hits, pivot moves).
\end{itemize}

These decisions indicate \emph{emergent tactical logic}: LLMs leveraged pre-trained general knowledge to navigate novel battle states without any hard-coded rules.

\subsubsection{Tournament Performance Summary}

The single-elimination bracket consisted of 8 competing models. The final standings are presented below.

\begin{table}[h!]
\centering
\begin{tabular}{|l|c|l|}
\hline
\textbf{Model} & \textbf{Record} & \textbf{Final Standing} \\ \hline
\texttt{o4-mini} & 3-0 & \textbf{Champion} \\ \hline
\texttt{o3} & 2-1 & Runner-up \\ \hline
\texttt{gpt-4.1} & 1-1 & Semi-finalist \\ \hline
\texttt{claude-sonnet-4} & 1-1 & Semi-finalist \\ \hline
\texttt{claude-3-5-sonnet} & 0-1 & Quarter-finalist \\ \hline
\texttt{claude-3-7-sonnet} & 0-1 & Quarter-finalist \\ \hline
\texttt{gemini-2.5-pro} & 0-1 & Quarter-finalist \\ \hline
\texttt{gemini-2.5-flash} & 0-1 & Quarter-finalist \\ \hline
\end{tabular}
\end{table}

\begin{figure}[h]
    \centering
    \includegraphics[width=1\linewidth]{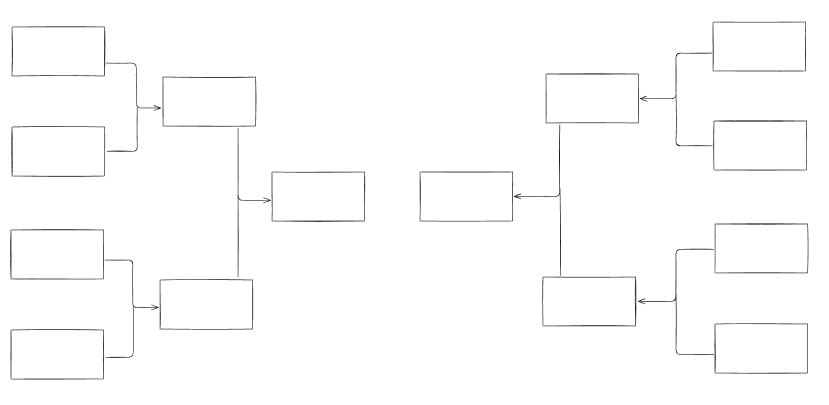}
    \caption{Complete single-elimination tournament bracket.}
    \label{fig:bracket}
\end{figure}

\subsubsection{Case Study: Championship Final – o4-mini vs. o3}

The final match pitted two distinct philosophies against each other:

\vspace{0.5em}
\noindent\textbf{o4-mini (Champion):} Kyogre, Groudon, Rayquaza, Lugia, Magnezone, Ho-Oh. \\
\textbf{o3 (Runner-up):} Swampert, Zapdos, Metagross, Blissey, Gengar, Salamence.
\vspace{0.5em}

\texttt{o3} followed a balanced, competitive team-building archetype that performed well throughout the bracket. However, \texttt{o4-mini}’s unique strategy of stacking high-stat legendaries with synergistic weather effects like rain for Kyogre and sun for Groudon/Ho-Oh—produced overwhelming offensive pressure. The opponent’s defensive core was unable to withstand the sheer force and tempo established by the champion team.

\subsubsection{Strategic Significance}

The tournament underscores that LLMs are capable of developing distinct strategic “personalities.” Most models converged on balanced team archetypes, which mirror human conventions in competitive play. In contrast, the winning model exploited an underutilized but valid strategy, prioritizing overwhelming stats and weather synergy which outperformed more conservative compositions. This diversity of approaches reflects the flexibility of LLMs in applying general knowledge creatively in adversarial, structured environments.


\begin{thebibliography}{00}
\bibitem{b1}Jia, Y., et al. (2025). Disentangling Reasoning Ability and Contextual Effects in Large Language Models via Behavioral Game Theory. \textit{Preprint}. 
\bibitem{b2} Lee, S., \& Kader, A. (2024). Evaluating Strategic Reasoning of Large Language Models in Behavioral Economics Games. \textit{AAAI Conference on Artificial Intelligence}. 
\bibitem{b3} Zhang, H., et al. (2025). K-Level Reasoning: Recursive Theory of Mind in Large Language Models. \textit{NeurIPS}. 
\bibitem{b4}Anthropic. (2025). How We Built Our Multi-Agent Research System. \textit{Anthropic Engineering Blog}. 
\bibitem{b5}Guan, X., et al. (2024). Deliberative Alignment: Enhancing Safety and Robustness in Large Language Models through Symbolic Reasoning and Reinforcement Learning. \textit{Technical Report}. 
\bibitem{b6} Yuan, X., et al. (2025). Tracing LLM Reasoning Processes with Strategic Games: A Framework for Planning, Revision, and Resource-Constrained Decision Making. \textit{arXiv preprint arXiv:2506.12012}. 
\bibitem{b7} Lv, Z., \& Qihang, C. (2024). Pokémon Battle Agent based on LLMs. \textit{THU Fall AML Submission} 
\bibitem{b8}Hu, S., Huang, T., Liu, G., Kompella, R. R., \& Liu, L. (2025). PokéLLMon: A Grounding and Reasoning Benchmark for Large Language Models in Adversarial Pokémon Battles. \textit{ICLR Workshop}. 
\bibitem{b9}VGC AI Competition. (2025). IEEE Conference on Games (CoG 2025) AI Competition. \href{https://cog2025.inesc-id.pt/vgc-ai-competition/}{https://cog2025.inesc-id.pt/vgc-ai-competition/} 

 

\end{thebibliography}
\end{document}